\documentclass[sigconf,natbib=true,anonymous=false]{acmart}
\renewcommand\footnotetextcopyrightpermission[1]{} 
\usepackage{listings}[language=Python]

\AtBeginDocument{%
  \providecommand\BibTeX{{%
    \normalfont B\kern-0.5em{\scshape i\kern-0.25em b}\kern-0.8em\TeX}}}

\begin{document}

\title{Low-variance estimation in the Plackett-Luce model via quasi-Monte Carlo sampling}

\author{Alexander Buchholz, Jan Malte Lichtenberg, Giuseppe Di Benedetto, Yannik Stein, Vito Bellini, Matteo Ruffini}
\email{{buchhola,jlichten,bgiusep,syannik,vitob,ruffinim}@amazon.com}
\affiliation{Amazon Music ML}

\renewcommand{\shortauthors}{Buchholz et al.}

\begin{abstract}
  The Plackett-Luce (PL) model is ubiquitous in learning-to-rank (LTR) because it provides a useful and intuitive probabilistic model for sampling ranked lists. 
  Counterfactual offline evaluation and optimization of ranking metrics are pivotal for using LTR methods in production. When adopting the PL model as a ranking policy, both tasks require the computation of expectations with respect to the model. These are usually approximated via Monte-Carlo (MC) sampling, since the combinatorial scaling in the number of items to be ranked makes their analytical computation intractable. Despite recent advances in improving the computational efficiency of the sampling process via the Gumbel top-k trick~\cite{Kool2019}, the MC estimates can suffer from high variance. We develop a novel approach to producing more sample-efficient estimators of expectations in the PL model by combining the Gumbel top-k trick with quasi-Monte Carlo (QMC) sampling, a well-established technique for variance reduction.
  We illustrate our findings both theoretically and empirically using real-world recommendation data from Amazon Music and the Yahoo learning-to-rank challenge.
\end{abstract}





\maketitle
\pagestyle{plain}

\section{Introduction}

In streaming media services with enormous catalogs, it is paramount to 
provide customers with personalized content, so that they can readily access 
media they are most likely to engage with. Often content is displayed in the
form of widgets, ranked from top to bottom in the home page, containing for instance songs from the same music genre or artist.
Many approaches to solve these ranking problems are based on the Plackett-Luce model \cite{luce1959individual, plackett1975analysis}, a probabilistic model for ranked lists, that given relevance scores, performs repeated sampling from a softmax distribution. Removing drawn items at each turn, the model constructs a list where items most likely to be sampled appear in top positions. 
As such it has been widely used in economics and statistics \cite{BEGGS19811, fok2012rank, koop1994rank, hausman1987specifying} as well as in recommender systems and information retrieval \cite{burges2010ranknet, singh2019policy, diaz2020evaluating}. Despite the intuitive way of sampling from the model, computing expectations  with respect to the PL  distribution is a challenging task. 
The computational burden comes from enumerating all possible combinations of list entries, and it leads to a complexity of $\mathcal{O}(k!)$ where $k$ is the size of the list. 
 Expectations with respect to the PL model are required, for example, when computing utilities of rankings and their gradients \cite{singh2019policy, oosterhuis2021computationally} or propensities, needed for counterfactual offline policy evaluation \cite{li2018offline}. 

Monte Carlo sampling allows to approximate resulting 
expectation but suffers from high variance of the resulting estimator. 
As a remedy to the mentioned problems we suggest the use of quasi-Monte Carlo (QMC) \cite{caflisch1998monte}, a well known variance reduction technique. 
Approximating expectations with QMC can lead to mean squared error rates of up to $\mathcal{O}(1/N^2)$ compared 
with the Monte Carlo (MC) rate of $\mathcal{O}(1/N)$ where $N$ is the number of samples. 

In what follows we will illustrate how the reduced variance that comes from QMC sampling leads to 
better estimation of propensities as well as more precise utilities and gradients in ranking models that are based on the PL model. We will combine our approach with the Gumbel top-k trick \cite{Kool2019,oosterhuis2021computationally} to generate efficient, low variance samples from the PL distribution. Our contributions consist in 
\begin{itemize}
  \item Introducing QMC to the field of learning-to-rank (LTR) and highlighting its ease of use in a wide range of settings;
  \item Deriving theoretical guarantees that show trade-offs between query batch sizes and Monte Carlo sample sizes when using QMC for LTR;
  \item Showing practical gains through experiments on the Yahoo LTR data \cite{chapelle2011yahoo} and production logs from Amazon Music. 
\end{itemize}

The rest of our work is structured as follows. In Section \ref{sec:related_work} we discuss how 
our contribution relates to recent work, Section \ref{sec:background} provides required background and sets the stage for our suggested approach in Section \ref{sec:qmc_for_pl}, Section \ref{sec:experiments} 
shows our experimental results and finally Section \ref{sec:conclusion} concludes.

\section{Related work} \label{sec:related_work}
\paragraph{Plackett-Luce model and learning-to-rank}
The PL model has seen growing attention in the machine learning community over recent years \cite{guiver2009bayesian, cheng2010label}. Used inside ranking models, the PL model leads to robust performance in industrial settings due to the probabilistic nature 
of the model \cite{bruch2020stochastic}. This is a quality that also leads to desirable exploration properties due to the implicit quantification of uncertainty \cite{oosterhuis2018differentiable, oosterhuis2021unifying}. 
However, efficiently training the ranking objective is a computationally difficult task \cite{electronics11010037}, in particular because policy gradients \cite{williams1992simple} can suffer from high variance, which consequently has lead to improved gradient computation procedures \cite{oosterhuis2021computationally, mohamed2020monte}. 

\paragraph{Quasi-Monte Carlo sampling for machine learning}
Quasi-Monte Carlo (QMC) is a variance reduction technique that has seen wide adoption in finance \cite{glasserman2004monte} and found its way into the statistics \cite{Buchholz2019a, letham2019constrained} and machine learning community \cite{Buchholz2018a, balandat2020botorch, liu2021quasi, wenzel2018quasi}. Its use can improve training of gradient based optimization methods that rely on sampling from a parametric distribution. 
To the best of our knowledge, we are not aware of any usage for learning-to-rank problems so far. 

\section{Background} \label{sec:background}
We recall here the building blocks for our suggested method. 

\subsection{Offline learning and evaluation using propensities}
In learning-to-rank a common objective is to find the best policy $\pi^\star$ that maximizes expected utility in a policy class $\pi \in \Pi$ 
\begin{eqnarray} \label{eq:rankingobjective}
  \pi^\star = \arg \max_{\pi \in \Pi} \mathbb{E}_{q \sim Q}[U(\pi | q)], 
\end{eqnarray}
where queries $q$ (list of items to rank) are sampled from a query distribution $Q$. The utility of a policy given a query is defined as 
\begin{eqnarray} \label{eq:utility}
  U(\pi | q) = \mathbb{E}_{r \sim \pi(r|q)}[\Delta(r, \text{rel}^q)],
\end{eqnarray}
where $r$ denotes a ranked list, $\text{rel}^q$ denotes the relevance of the items in query $q$ and $\Delta(\cdot, \cdot)$ is a ranking loss such as DCG or NDCG. See \cite{singh2019policy} for more details. 
In practical settings, we often only dispose of data coming from a different \textit{logging policy} $h$ and hence off-policy learning and evaluation methods are needed \cite{li2018offline}. We rewrite \eqref{eq:utility} as 
\begin{eqnarray} \label{eq:ipsutility}
  U(\pi | q) = \mathbb{E}_{r \sim h(r|q)}\left[\Delta(r, \text{rel}^q) \frac{\pi(r|q)}{h(r|q)}\right],
\end{eqnarray}
using an importance sampling identity, see \cite{Joachims2018, agarwal2019general, Swaminathan2015a} for more details. Depending on the assumptions made on the underlying click models we either use $h(r|q)$ directly or the propensity that item $i \in r$ is shown in position $k$, denoted by $h(i, k|q) = \mathbb{P}(\text{rank}(i)=k)$. If logged propensities are not available, they can be approximated using Monte Carlo using sampling from the policy $h$ in order to get 
\begin{eqnarray} \label{eq:prop_estimation}
  \mathbb{P}(\text{rank}(i)=k) \approx \sum_{n=1}^N 1{\{\text{rank}(i_n)=k\}}/N.
\end{eqnarray}
Then, $\widehat{h} \approx h$ is used inside \eqref{eq:ipsutility}, see also \cite{Ai2018}. Policy evaluation or training can be done on these approximate utilities. 

\subsection{Plackett-Luce sampling and the policy-gradient ranking algorithm}
A natural policy over rankings is the Plackett-Luce policy $\pi_\theta$, defined as a product of softmax distributions with scores $s_{i} = h_\theta(x_{i}^q)$, parametrized by a scoring function $h_\theta$ (e.g. neural network), which takes query-item features $x_{i}^q$ as input.
Optimizing the policy is achieved by finding 
$\theta^\star = \arg \max_{\theta} \mathbb{E}_{q \sim Q}[U(\pi_\theta | q)]$, typically by using gradient-based methods. 
However, differentiating the previous expression is difficult as the utility is itself an integral with respect to the policy $\pi_\theta$. The log derivative trick provides a solution \cite{williams1992simple}. 
The gradient of the utility can be approximated as
\begin{eqnarray} \label{eq:scorefunctiongradient}
  \nabla_\theta \mathbb{E}_{q \sim Q}[U(\pi_\theta | q)] \approx \frac{1}{QN} \sum_{q=1}^Q \sum_{n=1}^N \nabla_\theta \log \pi_\theta(r_{n,q} | q) \Delta(r_{n,q}, \text{rel}^q),
\end{eqnarray}
where for every query $q$ we draw $N$ rankings from the Plackett-Luce policy $\pi_\theta( \cdot | q)$. Estimated gradients are then used inside well established stochastic gradient descent methods where we use query batches of size $Q$ and $N$ samples from the PL model per query. 

This framework has two potential issues:
first, the score function gradient typically has high variance, which can lead to slow training \cite{ranganath2014black}. 
Second, sampling from the PL model can become costly for huge query sizes. 
We will show an efficient solution consisting in the combination of QMC with the Gumbel top-k trick \cite{Kool2019}.

\subsection{Sampling from the PL-model and the Gumbel top-k trick}
Sampling from the PL distribution can be achieved through repeated softmax sampling where the last selected item is removed from the list. This naive approach has complexity $\mathcal{O}(n_q^2)$ due to the repeated normalization of the softmax.
The Gumbel top-k trick allows to sample from the PL distribution  with log-linear complexity and was popularized by \cite{Kool2019} and recently introduced to the ranking literature by \cite{oosterhuis2021computationally}. 
A sample $r$ can be obtained by first sampling $u_1, \dots, u_{|q|} \sim \mathcal{U}[0,1]$, followed by computing standard Gumbel random variables $g_i = - \log(-\log(u_i)) \forall i$ and finally sorting the Gumbel variables summed with the scores. We obtain $r = \arg \text{sort}_i~(s_i + g_i)$ and thus the required $r \sim \pi(\cdot|q)$.

\subsection{Quasi-Monte Carlo}
Quasi-Monte Carlo \cite{caflisch1998monte,glasserman2004monte} is a numeric integration technique that has seen extensive
use in statistics and machine learning over the last years. Analogously to 
Monte Carlo sampling the goal is to approximate the integral 
$\int_{[0,1]^d} f(u) d u = I$.
Standard Monte Carlo sampling achieves this by formulating the above 
problem as
integral $I = \mathbb{E}[f(U)]$, where $U \sim \mathcal{U}[0,1]^d$. 
By the law of large numbers $I_N = \sum_{n=1}^N f(u_n)/N$, 
where $u_n \sim \mathcal{U}[0,1]^d$ for $n \in 1, \dots, N$,
converges to $I$ as $N \rightarrow \infty$. This convergence happens at a rate of
$\text{Var}(I_N) = \mathcal{O}(1/N) $. Going beyond the uniform distribution, the above
approach can be applied to almost arbitrary distributions $X \sim \mathbb{P}$ by exploiting the fact that
$X = \Gamma(U)$ and $\Gamma$ is the inverse cdf of $X$
and hence $I_N = \sum_{n=1}^N f(\Gamma(u_n))/N$ approximates $\mathbb{E}[f(X)]$. 

\begin{figure}[h] 
  \centering
  \includegraphics[scale=0.5]{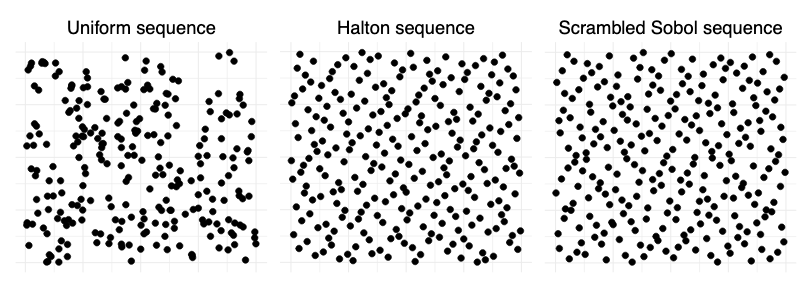}
  \caption{256 samples generated by a MC sequence (left), and two popular QMC sequences (middle and right) over $[0,1]^2$.}
    \label{fig:mcsequences}
\end{figure}

Quasi-Monte Carlo builds on the above by constructing low discrepancy sequences over the unit hypercube $u_1, \dots, u_N \in [0,1]^d$. These sequences can be thought of as being 
more evenly distributed than random uniform sampling. Under smoothness conditions on the 
function of interest $f \circ \Gamma$, quasi-Monte Carlo then achieves better approximations of $I_N$: the error behaves as $\mathbb{E}[|I-I_N|^2] = o(1/N)$ and thus goes to $0$ at a rate faster than standard MC, see for example \cite{gerber2015integration}. This rate can be as fast 
as $\mathcal{O}(1/N^2)$, see \cite{gerber2015integration}, and empirically almost always leads to a smaller error. 
We illustrate three different uniform sequences in Figure \ref{fig:mcsequences}. The left sequence corresponds to a random uniform Monte Carlo sequence and shows some typical random clustering. The middle (called Halton sequence \cite{halton1964algorithm}) and right sequences (called scrambled Sobol sequence \cite{sobol1967distribution, dick2011higher}) are more evenly distributed, hence using them for integration will result in a reduced error. 
In practice we use randomized QMC sequences, such as the scrambled Sobol sequence, which reintroduce
stochasticity in the sequence but keep the desirable properties.
Thus, expectations are well defined, the estimates are unbiased and we can use our probabilistic toolbox to compute variances. 
We refer to \cite{glasserman2004monte, dick2010digital} for more details on construction of the sequences.

\paragraph{When not to use quasi-Monte Carlo instead of regular MC} Despite the highlighted advantages, there are minor caveats to consider. First, QMC can suffer from a curse of dimensionality. If we integrate over high dimensional subspaces, we need more points to construct evenly spread sequences. Second, the function of interest to integrate $f \circ \Gamma$ must be sufficiently smooth, i.e., the function must at least twice integrable (the variance must exist). Third, sample sizes have to be set to $2^k$ for $k \in \mathbb{N}$ in order to obtain theoretical guarantees, \cite{owen2020dropping}. In practice, however, this is just a minor caveat. 

\section{QMC for PL sampling} \label{sec:qmc_for_pl}
We suggest to leverage the variance reduction from QMC for two purposes: (i) for obtaining low variance propensity estimates; (ii) for obtaining more precise gradients 
for our ranking objective in \eqref{eq:rankingobjective}. The combination of these ideas leads to a straightforward, easy-to-implement (see Listing \ref{lst:code}) variance reduction as QMC generators are available in widely used libraries such as \texttt{scipy} \cite{2020SciPy-NMeth} and \texttt{pytorch} \cite{NEURIPS2019_9015}. 
In summary we advocate to use the Gumbel top-k trick to sample $N$ rankings for every query $q$ by generating a QMC sequence of length $N$ and dimension $|q|$. Estimators of the utility in \eqref{eq:utility} will thus be of reduced variance and consequently the 
estimated gradients in \eqref{eq:scorefunctiongradient} will also be more precise. 
Our approach can also be used for estimating propensities that are required for counterfactual offline evaluation of ranking algorithms \cite{li2018offline}. 
Here is a code example to show how straightforward the implementation is:

\lstdefinestyle{CustomStyle}{
  language=Python,
  stepnumber=1,
  numbersep=10pt,
  tabsize=4,
  showspaces=false,
  showstringspaces=false
}
\lstset{basicstyle=\footnotesize, style=CustomStyle}

\begin{lstlisting}[caption={\mbox{Sample rankings from PL policy with MC and QMC}}, label={lst:code},captionpos=b]
import numpy as np
from torch.quasirandom import SobolEngine

def mc_rank_sampling(mc_type, mc_samples, n_actions, scores):
  if mc_type == "MC":
    u = np.random.uniform(size=(mc_samples, n_actions))
  elif mc_type == "QMC":
    s = SobolEngine(dimension=n_actions, scramble=True)
    u = s.draw(mc_samples).T.numpy()
  gumbels = -np.log(-np.log(u))
  return np.argsort(scores + gumbels, axis=1)
\end{lstlisting}

\subsection{Theoretical considerations}
As gets evident from \eqref{eq:scorefunctiongradient}, the gradient is an estimator that suffers from two sources of variability: the variance from the mini batch size of the queries (using $Q$ both for the number of queries and the query distribution) and the variance from sampling from the PL-model (with $N$ samples per query). 
As such, both terms will impact the precision of the estimators, a situation that has been studied in a reverse setting by \cite{Buchholz2019a}. 
Using a similar variance decomposition we obtain the following result.
 
\vspace{1cm}

\begin{theorem}
  Let $\hat{F}_{Q,N}= \frac{1}{QN} \sum_{q=1}^Q \sum_{n=1}^N \Delta(r_{n,q}, \text{rel}^q)$ be an estimator of $\mathbb{E}_{q \sim Q}[U(\pi_\theta | q)]$. If $r_{n,q}$ is based on samples of a random MC distribution, then 
  \begin{eqnarray}
    \text{Var}_Q[ \mathbb{E}_N [\hat{F}_{Q,N}| Q]] = \mathcal{O}\left(\frac{1}{Q}\right), \label{eq:slowpart} \\
    \mathbb{E}_Q[ \text{Var}_N [\hat{F}_{Q,N}|Q]] = \mathcal{O}\left(\frac{1}{QN}\right) \label{eq:mcpart}.
  \end{eqnarray}
  If $r_{n,q}$ is based on samples of a transformed QMC sequence (in particular a scrambled Sobol sequence) then 
  \begin{eqnarray}
    \mathbb{E}_Q[ \text{Var}_N [\hat{F}_{Q,N}|Q]] = \mathcal{O}\left(\frac{1}{QN^2}\right), \label{eq:qmcpart}
  \end{eqnarray}
  and the first term \eqref{eq:slowpart} stays the same. 
\end{theorem}
The proof is based on a decomposition of variance (overloading notation of $Q$ and $N$) using the law of total variance $$\text{Var}[\hat{F}_{Q,N}]  = \mathbb{E}_Q[ \text{Var}_N [\hat{F}_{Q,N}|Q]]+\text{Var}_Q[ \mathbb{E}_N [\hat{F}_{Q,N}| Q]] .$$
We exploit that both randomized QMC and MC yield unbiased estimators per query.
Then, we check that the transformation mapping $u_n$ to $\Delta(r_{n,q}, \text{rel}^q)$ is square-integrable. This is true as the transformation is bounded due to the Gumbel top-k trick. 
A similar result can be obtained for the gradient in \eqref{eq:scorefunctiongradient}, where we would use the squared norm of the gradient (omitted here due to space constraints). 
As shown in \cite{Buchholz2018a} (without the mini-batch perspective), the resulting stochastic gradient procedure results in faster convergence of the loss function optimization. 
Our result highlights two points: first, QMC will almost always result in more precise estimators due to the improved rate in \eqref{eq:qmcpart} compared to \eqref{eq:mcpart}. However, second, a natural limit exists as the dominating term in the decomposition is the batch size in \eqref{eq:slowpart}. The batch size variance contribution remains even if all variance from the PL sampling is removed. 
In our experiment section we will highlight these two points. 

\section{Experiments} \label{sec:experiments}
We illustrate how the obtained variance reduction from QMC bring improvements in two use cases. First, increased precision when computing propensities useful for counterfactual offline evaluation, and second, we will show performance improvement when training PG-rank due to reduced gradient variance using industrial production logs from a ranking use case in Amazon Music, as well as data from the Yahoo LTR challenge. 

\subsection{Propensity estimation}
\begin{figure}[h] 
  \centering
  \includegraphics[width=0.9\linewidth]{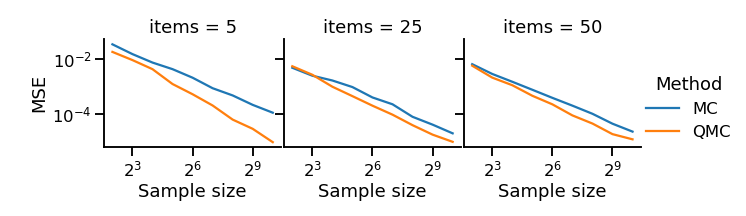}
  \caption{Propensity estimation for the PL model: MSE of MC and QMC estimates for different sample and list sizes.
  }
  \label{fig:propensity_estimation}
\end{figure}

We illustrate the improved estimation of propensities in \eqref{eq:prop_estimation} using QMC for the PL model. 
We fix lists of size $[5, 25, 50]$ with item scores generated according to a standard normal distribution $\mathcal{N}(0,1)$ with a fixed seed. We simulate draws from the PL model for sample sizes going from $2^2$ to $2^{10}$ to compute estimators of the propensities. We repeat this $200$ times to assess the variance of the the estimator in \eqref{eq:prop_estimation}. We illustrate the reduced error in estimating the propensity in Figure \ref{fig:propensity_estimation}. As shown, the QMC based estimator is consistently more precise (lower mean squared error) for different list sizes (number of items) and the error decreases faster than the equivalent estimator based on MC. As the list sizes increases from $5$ to $50$ the achieved MSE reduction decreases. 

\subsection{Training of PG-rank}
\begin{figure}[h] 
  \centering
  \includegraphics[width=0.95\linewidth]{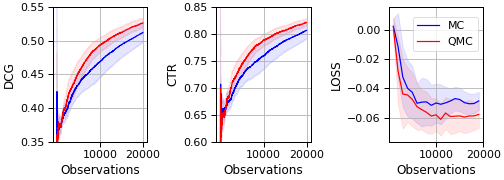}
  \caption{DCG, CTR, and loss when online training PG-rank on the Yahoo Learning to Rank challenge data. 
  }
  \label{fig:yltr}
\end{figure}

We illustrate improved training of PG rank using data from the YLTR challenge \cite{chapelle2011yahoo} and logs of a production policy used for a ranking use case in Amazon Music. 
\paragraph{YLTR}
We simulate an online learning framework by using the approach described in \cite{agarwal2019estimating, mayor2021ranker} using the position bias model for simulating if a relevant item was seen. We use $100$ different items to rank and set the position bias curve to $1/k$ where $k$ is the item position. We run over a single epoch of the dataset only, using a batch size of $1000$. The metrics are computed out-of-sample, meaning for every batch we computed loss, DCG and CTR before updating the model with the data from the batch, mimicking the production scenario where the model is trained online with batch updates. We use a neural network with 64 hidden units inside PG rank. 
The algorithms runs over one epoch of data using SGD with a learning 
rate of $0.1$. We repeat the run 10 times to obtain confidence intervals. 
As illustrates Figure \ref{fig:yltr}, the use of as little as $8$ quasi-Monte Carlo samples speeds up learning substantially. We see an increase of the DCG, the CTR and a reduction in the loss as learning progresses. 

\paragraph{Real world production logs}
Finally, we illustrate the applicability of our approach on production 
data from Amazon Music.
The dataset contains roughly $10^6$ data points that were logged from a deterministic ranking policy. Our offline training approach uses a position bias model to correct for the fact that implicit feedback of the user depends on the visibility of the content. The position bias curve was estimated 
using the approach described in \cite{agarwal2019estimating}. We use the offline evaluation methodology as outlined in \cite{li2018offline}. We train PG-rank algorithms using SGD with different step sizes using a single layer neural network of size 64 and with a learning rate of $0.001$. We vary query batch sizes and Monte Carlo sampling sizes in order to disentangle their respective effect. 

To avoid disclosing sensitive business information we report relative improvements (lift in \% of the target metrics) of QMC with respect to the MC version of the algorithm only. We repeat estimations 10 times in order to obtain confidence intervals. 

\begin{figure}[h] 
  \centering
  \includegraphics[scale=0.3]{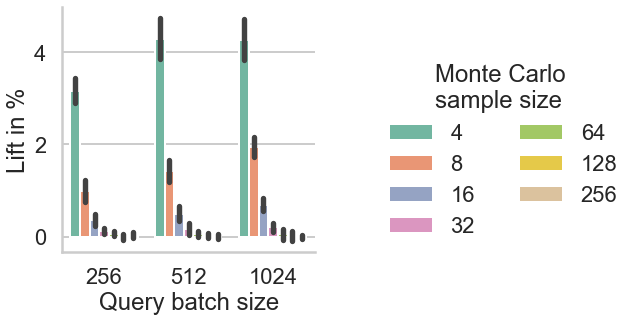}
  \caption{Lift in \%  of our target metric when using QMC instead of MC 
  for training the PG-rank algorithm on production logs of Amazon Music. 
  }
  \label{fig:qmc_improvement}
\end{figure}

As illustrated in Figure \ref{fig:qmc_improvement}, the use of QMC instead of MC sampling leads to 
improvements in the training objective of up to $4\%$. 
At worst, there is no notable difference between the two algorithms. The gain decreases when we allow for more Monte-Carlo samples, which in turn increases the computational cost of the model updates.
Importantly, there was no notable difference in computational time between training using MC and QMC.

\section{Conclusion} \label{sec:conclusion}
We showed that QMC is a straightforward approach for sampling from the PL model.
QMC yields low-variance, unbiased estimators of propensities
and speeds up training of learning-to-rank methods 
as highlighted with PG-rank. 
This provides a strong case for using QMC consistently 
in production models as the implementation is easy and comes with guarantees. 
Potential extensions are the comparison and generalization to other ranking algorithms 
that use sampling from the PL model to approximate loss functions or their gradients such as \cite{47258} or \cite{oosterhuis2021computationally} and a more thorough investigation of the offline learning setting. We conjecture that in these models QMC will also lead to convergence improvements. 
\bibliographystyle{ACM-Reference-Format}
\bibliography{sample-base}










\end{document}